\documentclass[letterpaper, 10 pt, conference]{ieeeconf}

\IEEEoverridecommandlockouts
\usepackage[T1]{fontenc}
\usepackage{amsmath}
\usepackage{graphicx}
\usepackage[ruled,linesnumbered]{algorithm2e}
\usepackage{booktabs}
\usepackage{multirow}
\usepackage{balance}
\usepackage[hidelinks=true,bookmarks=FWLse]{hyperref}
\usepackage{caption}
\usepackage{lineno}
\usepackage{comment}
\usepackage{color}
\usepackage{cite}
\usepackage{algorithmic}
\usepackage{tabularx}
\usepackage{xcolor}
\usepackage{mathrsfs}

\usepackage{url,subfigure,multirow,tabulary}

\newcommand{\ie}{\textit{i.e.}}
\newcommand{\eg}{\textit{e.g.}}

\usepackage{array}
\makeatletter
\newcommand{\thickhline}{%
    \noalign {\ifnum 0=`}\fi \hrule height 1pt
    \futurelet \reserved@a \@xhline
}
\newcolumntype{"}{@{\hskip\tabcolsep\vrule width 1pt\hskip\tabcolsep}}
\makeatother

\makeatletter

\newcommand{\Rmnum}[1]{\expandafter\@slowromancap\romannumeral #1@}
\makeatother

\newcolumntype{L}[1]{>{\raggedright\arraybackslash}p{#1}}
\newcolumntype{C}[1]{>{\centering\arraybackslash}p{#1}}
\newcolumntype{R}[1]{>{\raggedleft\arraybackslash}p{#1}}

\title{\LARGE \bf
SNE-RoadSeg+: Rethinking Depth-Normal Translation and\\Deep Supervision for Freespace Detection
}

\author{Hengli Wang$^1$*, Rui Fan$^2$*, Peide Cai$^1$, and Ming Liu$^1$, \IEEEmembership{Senior Member, IEEE}
\thanks{\textit{(Corresponding author: Ming Liu.)}}
\thanks{$^1$H. Wang, P. Cai and M. Liu are with the Department of Electronic and Computer Engineering, the Hong Kong University of Science and Technology, Clear Water Bay, Kowloon, Hong Kong SAR, China (email: \{hwangdf, peide.cai, eelium\}@ust.hk).}
\thanks{$^2$R. Fan is with the College of Electronic and Information Engineering, Tongji University, Shanghai 201804, P. R. China, as well as Shanghai Research Institute for Intelligent Autonomous Systems, Shanghai 201210, P. R. China (email: rui.fan@ieee.org).}
\thanks{*The authors contributed equally to this work.}
}

\begin{document}

\maketitle
\thispagestyle{empty}
\pagestyle{empty}

\begin{abstract}
Freespace detection is a fundamental component of autonomous driving perception. Recently, deep convolutional neural networks (DCNNs) have achieved impressive performance for this task. In particular, SNE-RoadSeg, our previously proposed method based on a surface normal estimator (SNE) and a data-fusion DCNN (RoadSeg), has achieved impressive performance in freespace detection. However, SNE-RoadSeg is computationally intensive, and it is difficult to execute in real time. To address this problem, we introduce SNE-RoadSeg+, an upgraded version of SNE-RoadSeg. SNE-RoadSeg+ consists of 1) SNE+, a module for more accurate surface normal estimation, and 2) RoadSeg+, a data-fusion DCNN that can greatly minimize the trade-off between accuracy and efficiency with the use of deep supervision. Extensive experimental results have demonstrated the effectiveness of our SNE+ for surface normal estimation and the superior performance of our SNE-RoadSeg+ over all other freespace detection approaches. Specifically, our SNE-RoadSeg+ runs in real time, and meanwhile, achieves the state-of-the-art performance on the KITTI road benchmark. Our project page is at \url{https://www.sne-roadseg.site/sne-roadseg-plus}.
\end{abstract}

\section{Introduction}
\label{sec.introduction}
Autonomous driving appears prominently in our society in the form of the advanced driver assistance system (ADAS) in both commercial and research vehicles \cite{van2018autonomous}. Visual environment perception, the front-end module and key component of the ADAS, analyzes the raw data collected by the car's sensors and outputs its understanding to the driving scenario \cite{yan2017supervised,wang2021pvstereo,wang2020cot}. Its outputs are then used by other modules, such as prediction and planning, to ensure the safe navigation of self-driving cars in complex environments \cite{wang2021learning,wang2021end}.

As a fundamental task in visual environment perception, freespace detection performs pixel-level binary classification on vision sensor data, \eg, RGB images \cite{han2018semisupervised}, LiDAR point clouds \cite{chen2019progressive}, or depth/disparity images \cite{fan2019road}. This is generally realized with traditional segmentation algorithms and/or deep convolutional neural networks (DCNNs) \cite{fan2021learning}. With the use of modern encoder-decoder architectures, semantic segmentation DCNNs have emerged as the most powerful tool for robust freespace detection, and their performance under different environmental conditions is incredibly good. Therefore, many researchers have turned their focuses towards developing DCNN-based freespace detection approaches.

Recent data-fusion DCNNs for semantic segmentation \cite{wang2019self,fan2020sne, wang2020applying, wang2021dynamic} have achieved the state-of-the-art (SOTA) performance in freespace detection by extracting visual features from different modalities of vision sensor data and fusing the extracted features to provide accurate semantic prediction. For example, progressive LiDAR adaptation-aided road detection (PLARD) \cite{chen2019progressive} learns both visual and LiDAR features using two DCNNs. A feature space adaptation module then follows to adapt the LiDAR features to visual features. This helps PLARD \cite{chen2019progressive} achieve impressive freespace detection results. Furthermore, we recently introduced a SOTA freespace detection algorithm, named SNE-RoadSeg~\cite{fan2020sne}. It consists of 1) a surface normal estimator (SNE), a lightweight module for efficient end-to-end translation from depth/disparity images into surface normal inference maps, and 2) RoadSeg, a data-fusion DCNN capable of extracting and fusing features from both RGB images and the inferred surface normal maps for accurate freespace detection. However, SNE-RoadSeg~\cite{fan2020sne} is computationally intensive, and it is difficult to execute in real time. One possible solution is to reduce the network depth/level of RoadSeg, but the selection of the optimal depth requires an extensive architecture search. Furthermore, in order to improve the computational efficiency, the SNE hypothesizes that the angle between an arbitrary pair of normalized surface normals is less than $\pi/2$ \cite{fan2020sne}. This can, sometimes, degrade the performance of the SNE near ambiguities, further deteriorating freespace detection performance.

\begin{figure*}[t]
    \centering
    \includegraphics[width=0.99\textwidth]{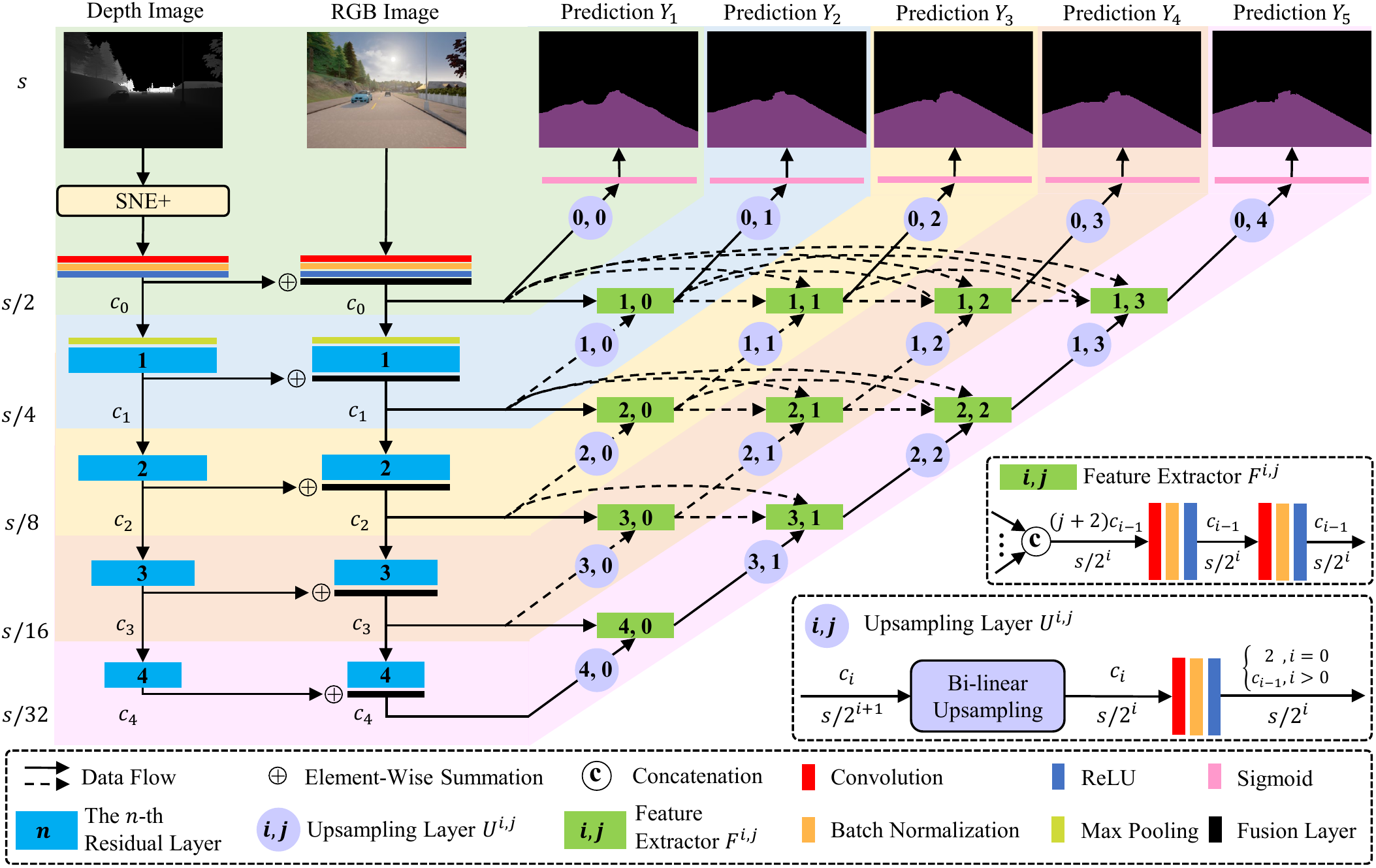}
    \caption{An overview of our proposed freespace detection framework SNE-RoadSeg+. It consists of 1) SNE+, a lightweight module for accurate surface normal information inference, and 2) RoadSeg+, a data-fusion DCNN that can greatly minimize the trade-off between accuracy and efficiency with the use of deep supervision. The entire framework has five network levels, which are illustrated in different background colors. During the inference phase, we can either 1) assemble all five freespace predictions to achieve the most accurate performance or 2) prune the network and only use the low-level parts to achieve the required efficiency without additional training. $s$ denotes the resolution of the input RGB and depth images, and $c_n$ denotes the number of feature map channels at different levels.}
    \label{fig.roadseg+}
\end{figure*}

To resolve the limitations of SNE-RoadSeg, in this paper, we introduce SNE-RoadSeg+, an upgraded version of SNE-RoadSeg, as shown in Fig.~\ref{fig.roadseg+}. SNE-RoadSeg+ consists of 1) SNE+, a module for surface normal information inference without the hypothesis made in \cite{fan2020sne}, and 2) RoadSeg+, a data-fusion DCNN that can greatly minimize the trade-off between accuracy and efficiency with the use of deep supervision. Extensive experimental results on the DIODE \cite{diode_dataset} and ScanNet \cite{dai2017scannet} datasets have demonstrated the effectiveness of our SNE+ for surface normal inference. Moreover, we have evaluated our SNE-RoadSeg+ on the KITTI road \cite{fritsch2013new} and Ready-to-Drive (R2D) road \cite{fan2020sne} datasets. The achieved results show that our SNE-RoadSeg+ outperforms all other freespace detection approaches. Specifically, SNE-RoadSeg+ runs in real time, and meanwhile, achieves the state-of-the-art performance on the KITTI road benchmark\footnote{\url{www.cvlibs.net/datasets/kitti/eval_road.php}} \cite{fritsch2013new}.

\section{Related Work}
\label{sec.related_work}

\subsection{Surface Normal Estimation}
Existing surface normal estimation methods can be basically categorized as data-driven \cite{bansal2016marr,huang2019framenet} or geometry-based \cite{hinterstoisser2011gradient,jordan2014quantitative}.
The former algorithms are generally trained using supervised learning techniques, hence requiring a large amount of well-labeled surface normal ground truth to find the best network parameters. Moreover, such algorithms are not designed specifically for surface normal estimation, because they are only considered as auxiliary functionalities for other computer vision and robotics tasks. Prior to our works \cite{fan2021three,fan2020sne}, geometry-based algorithms typically fit local (planar or quadratic) surfaces to a collection of 3D  points adjacent to the observed 3D point. By minimizing residual error using optimization techniques, such as principal component analysis or singular value decomposition, the surface normal information can be obtained. Recently, we proposed three-filters-to-normal (3F2N) \cite{fan2021three} and SNE-RoadSeg \cite{fan2020sne}, which provide an efficient way to translate depth/disparity images directly into accurate surface normal maps. In this paper, we introduce SNE+, a more accurate surface normal estimation approach, which neglects the hypothesis made in SNE-RoadSeg \cite{fan2020sne}. In addition, our SNE+ can also benefit DCNNs in freespace detection, as will be demonstrated in Section~\ref{sec.experiments}.

\subsection{Semantic Segmentation DCNNs}
Since \cite{long2015fully} proposed the fully convolutional network (FCN), many single-modal networks have been developed for semantic segmentation. SegNet \cite{badrinarayanan2017segnet} first presented the popular encoder-decoder architecture, which is widely used in current networks. U-Net \cite{ronneberger2015u} adopts the encoder-decoder paradigm, and further incorporates skip connections into the network for performance improvement. Moreover, DeepLabv3+ \cite{chen2018encoder} and DenseASPP \cite{yang2018denseaspp} combine the advantages of the spatial pyramid pooling (SPP) module and the encoder-decoder architecture to improve the semantic prediction detail. In addition, GSCNN \cite{takikawa2019gated} utilizes the boundary information to refine the semantic predictions.

To further improve the semantic segmentation performance, some researchers have developed data-fusion networks that use two (or more) types of visual features. Specifically, FuseNet \cite{hazirbas2016fusenet} utilizes RGB images and depth images based on the popular encoder-decoder architecture. Similarly, Depth-aware CNN \cite{wang2018depth} presents two novel operations, depth-aware convolution and depth-aware average pooling, to extract useful information from depth images for performance improvement. Moreover, RTFNet \cite{sun2019rtfnet} was designed to perform semantic segmentation using RGB images and thermal images. Additionally, our SNE-RoadSeg~\cite{fan2020sne} first transforms the depth/disparity images into surface normal maps, and then fuses the features learned from both RGB images and the inferred surface normal maps for accurate freespace detection. However, SNE-RoadSeg \cite{fan2020sne} is computationally intensive, and it is difficult to execute in real time. SNE-RoadSeg+ is, thus, proposed to address this problem, and it can greatly minimize the trade-off between accuracy and efficiency with the use of deep supervision.

\subsection{Deep Supervision}
Deep supervision aims at improving the network performance by providing supervision on the intermediate layers of the network \cite{lee2015deeply}. This paradigm has been used in many tasks, such as semantic segmentation. For example, the architecture design of DenseASPP \cite{yang2018denseaspp} adopts the concept of deep supervision implicitly, while \cite{zhou2019unet++} added additional supervision layers to improve the performance of semantic segmentation. Following these studies, we also incorporate deep supervision into SNE-RoadSeg+ for accurate and efficient freespace detection, making SNE-RoadSeg+ the first data-fusion DCNN to adopt the deep supervision paradigm. Moreover, we follow \cite{zhou2019unet++} and adopt a model pruning approach to achieve a great trade-off between accuracy and efficiency using deep supervision.

\section{SNE+}
\label{sec.sne+}
We demonstrated in \cite{fan2020sne} and \cite{fan2021three} that the surface normal information can be accurately and efficiently inferred from dense depth/disparity images in an end-to-end manner. These approaches first estimate the gradient $\mathbf{g}_{xy}=(n_x;n_y)$ of the surface normal's projection on the $xy$-plane by  performing gradient filtering on a given disparity (or inverse depth) image \cite{fan2020sne}. The preliminaries of $\mathbf{g}_{xy}$ estimation are given in the supplement. Given a 3D point $\mathbf{p}$ adjacent to the observed 3D point $\mathbf{q}$, an $n_z$ candidate can be obtained. As all candidates share one $\mathbf{g}_{xy}$, their provided surface normals are on the same tangent spherical surface. Therefore, the estimation of $\hat{{n}}$ (the optimum ${{n}}$) can be realized by finding a point on the arc of this tangent spherical surface where the $n_z$ candidates are distributed most intensively. The key to designing an SNE thus turns into the way of formulating
\begin{equation}
    \hat{n}_z = \Phi(\mathbf{g}_{xy},\mathbf{q},\mathscr{P}),
    \label{eq.function}
\end{equation}
where $\mathscr{P}=(\mathbf{p}_1;\mathbf{p}_2;\dots;\mathbf{p}_k)$ is a group of $k$ neighboring 3D points around the observed 3D point $\mathbf{q}$. \cite{fan2021three} formulates (\ref{eq.function}) as a median or mean filter. The former achieves better accuracy, but meanwhile, it is more computationally intensive  because of the sorting operation \cite{fan2021three}. Furthermore, SNE-RoadSeg~\cite{fan2020sne} formulates (\ref{eq.function}) as an energy minimization problem with respect to inclination and azimuth. However, it hypothesizes that the angle between an arbitrary pair of surface normals is less than $\pi/2$, limiting its performance near ambiguities, where the surface normal candidates can differ significantly from each other. Since a surface normal is undirected ($\mathbf{n}$ and $-\mathbf{n}$ can be considered to be exactly the same), we formulate (\ref{eq.function}) as follows:
\begin{equation}
    \begin{aligned}
    \hat{n}_z &=\Big[\sin\theta\cos\varphi,\sin\theta\sin\varphi,\cos\theta\Big]^\top,
    \end{aligned}
    \label{eq.energy}
\end{equation}
where
\begin{equation}
    \begin{aligned}
        \theta &=
        \underset{\theta}
        {\arg\max} \sum_{i=1}^k(A_i \sin\theta + n_{z_i}\cos\theta)^2\\
        &=\frac{1}{2}\arctan
        \Bigg(
        \frac{2\sum_{i=1}^{k} A_i n_{z_i}}
        {\sum_{i=1}^{k} \left({n_{z_i}}^2 - {A_i}^2\right)}
        \Bigg) + \frac{\pi}{2}l, l \in \{0, 1\}
    \end{aligned}
    \label{eq.theta}
\end{equation}
$\in[0,\pi]$ denotes inclination, $\varphi\in[0,2\pi)$ denotes azimuth~\cite{fan2020sne}, and $A_i = n_{x_i}\cos\varphi  + n_{y_i} \sin\varphi$. Compared with the SNE used in SNE-RoadSeg~\cite{fan2020sne}, (\ref{eq.energy}) can produce a more accurate $\hat{n}_z$, and therefore, we refer to it as SNE+.

\section{RoadSeg+}
\label{sec.roadseg+}
Based on our previous work SNE-RoadSeg \cite{fan2020sne}, our SNE-RoadSeg+ first employs the above-mentioned SNE+ to translate depth/disparity images into surface normal maps, and then fuses the features learned from both RGB images and the inferred surface normal maps for accurate freespace detection, as shown in Fig.~\ref{fig.roadseg+}. The data-fusion DCNN follows the popular encoder-decoder paradigm. Specifically, the encoder first extracts and fuses the different modalities of features in a multi-scale fashion. The decoder then utilizes feature extractors $F^{i,j}$ and upsampling layers $U^{i,j}$ to realize flexible feature fusion and accurate freespace detection. Readers are recommended to refer to \cite{fan2020sne} for more details on the network architecture. The rest of this section mainly introduces the major difference between SNE-RoadSeg \cite{fan2020sne} and SNE-RoadSeg+, that is, incorporating deep supervision into the network to improve the accuracy and efficiency for freespace detection.

\begin{figure*}[t!]
    \centering
    \includegraphics[width=0.99\textwidth]{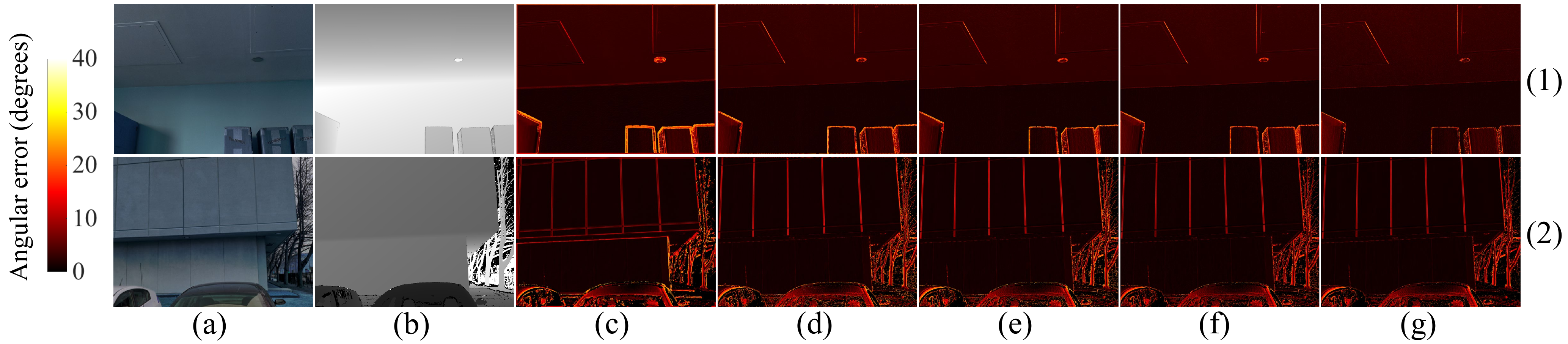}
    \caption{Examples of the surface normal estimation results on the DIODE dataset \cite{diode_dataset}: (a) RGB images; (b) depth images; (c)--(g) angular error maps of LINE-MOD \cite{hinterstoisser2011gradient}, PlanePCA \cite{jordan2014quantitative}, 3F2N \cite{fan2021three}, SNE \cite{fan2020sne} and our SNE+, respectively; (1) the indoor scenario; and (2) the outdoor scenario.}
    \label{fig.diode}
\end{figure*}

Different from SNE-RoadSeg \cite{fan2020sne}, which only uses one upsampling layer $U^{0,4}$ to perform the final freespace prediction, we append four upsampling layers, namely, $U^{0,0}$, $U^{0,1}$, $U^{0,2}$ and $U^{0,3}$, to output freespace predictions at different network levels/depths. During the training phase, the freespace prediction $Y_i$ at level $i$ is supervised by the cross entropy loss as follows:
\begin{equation}
    \mathcal{L}_{i} \left(\widehat{Y}, Y_{i}\right) = - \sum_{\mathbf{p}} \widehat{Y}\left(\mathbf{p}\right) \cdot \mathrm{log}\left( Y_i\left(\mathbf{p}\right) \right),
\end{equation}
where $\mathbf{p}$ denotes the valid pixels and $\widehat{Y}$ denotes the freespace ground truth. The adopted overall loss $\mathcal{L}$ is then defined as a weighted summation of $\mathcal{L}_i$, \ie, $\mathcal{L} = \sum_{i=1}^{5} \alpha_{i} \mathcal{L}_{i}$. During the inference phase, we can assemble the freespace estimations at all five network levels by computing their average to generate the final freespace prediction.

Additionally, we follow \cite{zhou2019unet++} and adopt a model pruning approach to optimize the trade-off between accuracy and efficiency based on the deep supervision paradigm. Specifically, since the freespace predictions at low network levels do not rely on the high-level network architecture, we can prune arbitrary high-level layers of the architecture to achieve flexible acceleration with acceptable performance degradation. For example, we can adopt the network with only two levels, \ie, the green and blue parts shown in Fig.~\ref{fig.roadseg+}, and the final freespace estimation can be obtained by assembling $Y_1$ and $Y_2$. Please note that this model pruning process can be directly conducted during the inference phase to achieve the required efficiency without additional training.

In summary, the advantages of the proposed deep supervision paradigm are twofold. 1) Providing supervision on the intermediate layers of the network can smooth the gradient flow for effective and efficient training, further leading to accurate freespace prediction results. 2) The deep supervision paradigm ensures that the intermediate layers of the network can provide accurate freespace predictions. Therefore, we do not require additional training after pruning the network. Instead, we can directly use the pruned network during the inference phase to boost network efficiency.

\section{Experimental Results and Discussion}
\label{sec.experiments}

\subsection{Datasets and Experimental Setups}
\label{sec.datasets}
Two datasets are adopted to evaluate the performance of our SNE+ for surface normal estimation:
\begin{itemize}
    \item The DIODE dataset \cite{diode_dataset}: This dataset provides approximately 25K depth images with the corresponding surface normal ground truth in real-world indoor and outdoor scenarios.
    \item The ScanNet dataset \cite{dai2017scannet}: This dataset contains about 2.5M depth images with the corresponding surface normal ground truth in 1513 real-world indoor scenarios.
\end{itemize}

Moreover, two other datasets are used to evaluate the performance of our SNE-RoadSeg+ for freespace detection:
\begin{itemize}
    \item The KITTI road dataset \cite{fritsch2013new}: This dataset provides RGB-D data in real-world driving scenarios. Specifically, it contains 289 paris of training data with ground truth for freespace detection and 290 pairs of testing data without ground truth.
    \item The R2D road dataset \cite{fan2020sne}: This dataset is a synthetic dataset collected under different illumination and weather conditions. It contains 11430 pairs of RGB-D images with the corresponding ground truth for freespace detection.
\end{itemize}

\begin{figure*}[t]
    \centering
    \includegraphics[width=0.99\textwidth]{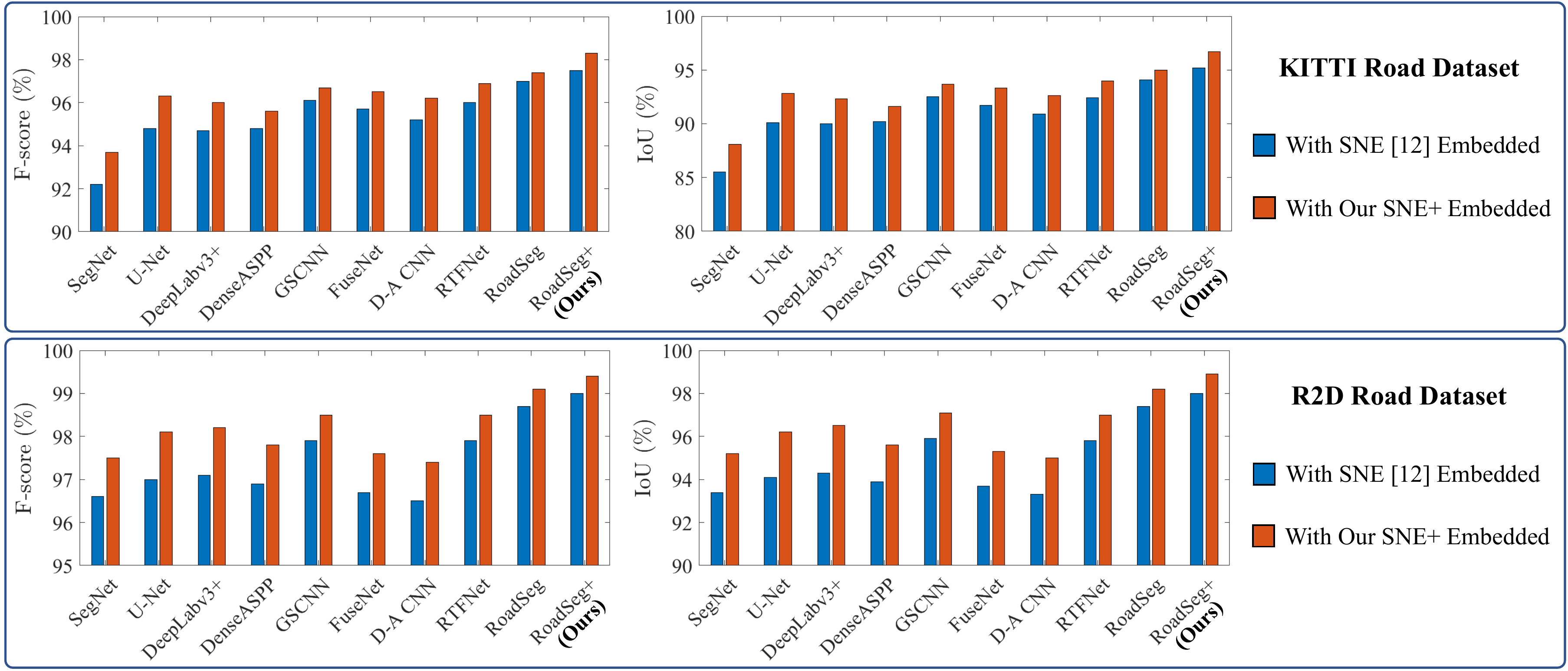}
    \caption{DCNN comparison for freespace detection performance on the KITTI road \cite{fritsch2013new} and R2D road \cite{fan2020sne} datasets. ``D-A CNN'' is the abbreviation of ``Depth-aware CNN''. SegNet \cite{badrinarayanan2017segnet}, U-Net \cite{ronneberger2015u}, DeepLabv3+ \cite{chen2018encoder}, DenseASPP \cite{yang2018denseaspp} and GSCNN \cite{takikawa2019gated} are single-modal DCNNs, while FuseNet \cite{hazirbas2016fusenet}, Depth-aware CNN \cite{wang2018depth}, RTFNet \cite{sun2019rtfnet}, RoadSeg \cite{fan2020sne} and our RoadSeg+ are data-fusion DCNNs.}
    \label{fig.road_evaluation}
    \vspace{-1em}
\end{figure*}

In our experiments, we first compare our SNE+ with state-of-the-art surface normal estimation approaches, as presented in Section~\ref{sec.surface_normal_evaluation}. Then, we compare our SNE-RoadSeg+ with state-of-the-art DCNNs for freespace detection. Specifically, we adopt the stochastic gradient descent with momentum (SGDM) optimizer during the training phase. We also adopt the early stopping mechanism to avoid over-fitting. The corresponding experimental results are presented in Section~\ref{sec.freespace_detection_evaluation}. Finally, we submit the results achieved by our SNE-RoadSeg+ to the KITTI road benchmark \cite{fritsch2013new}, as presented in Section~\ref{sec.benchmark}.

In addition, we adopt the average angular error $e_A$ to quantify the performance of surface normal estimation approaches:
\begin{equation}
    e_A = \frac{1}{m} \sum_{k=1}^{m} \cos ^{-1}\left(\frac{\left\langle\mathbf{n}_{k}, \hat{\mathbf{n}}_{k}\right\rangle}{\left\|\mathbf{n}_{k}\right\|_{2}\left\|\hat{\mathbf{n}}_{k}\right\|_{2}}\right),
\end{equation}
where $m$ denotes the number of valid (observed) pixels; and $\mathbf{n}_{k}$ and $\hat{\mathbf{n}}_{k}$ denote the ground-truth and estimated surface normals, respectively. An accurate surface normal estimation approach achieves a low $e_A$ value. Furthermore, two commonly used metrics are adopted for the performance evaluation of freespace detection, namely, the F-score (Fsc) and intersection over union (IoU):
\begin{align}
    \text{Fsc} &= \frac{2 n_{\mathrm{tp}}^{2}}{2 n_{\mathrm{t}_{\mathrm{p}}}^{2}+n_{\mathrm{tp}}\left(n_{\mathrm{fp}}+n_{\mathrm{fn}}\right)} \times 100\%,\\
    \text{IoU} &= \frac{n_{\mathrm{tp}}}{n_{\mathrm{tp}} + n_{\mathrm{fp}} + n_{\mathrm{fn}}} \times 100\%,
\end{align}
where $n_{\mathrm{tp}}$, $n_{\mathrm{tn}}$, $n_{\mathrm{fp}}$ and $n_{\mathrm{fn}}$ denote the true positive, true negative, false positive and false negative pixel numbers, respectively. An accurate freespace detection approach achieves high Fsc and IoU values.

\subsection{Evaluations for Surface Normal Estimation}
\label{sec.surface_normal_evaluation}
We compare our SNE+ with four SOTA surface normal estimation approaches: LINE-MOD \cite{hinterstoisser2011gradient}, PlanePCA \cite{jordan2014quantitative}, 3F2N \cite{fan2021three} and SNE \cite{fan2020sne}. The quantitative results are presented in Table~\ref{tab.normal}, where it can be seen that our SNE+ outperforms all other approaches on both the DIODE \cite{diode_dataset} and ScanNet \cite{dai2017scannet} datasets. Examples of the qualitative results are shown in Fig.~\ref{fig.diode}, where it can be observed that our SNE+ performs better near object boundaries. This is due to the ability of our proposed parameterization method to greatly minimize the effects caused by ambiguities.

\begin{table}[t]
    \caption{$e_A$ (degrees) of surface normal estimation approaches on the DIODE \cite{diode_dataset} and ScanNet \cite{dai2017scannet} datasets. The best results are shown in bold type.}
    \centering
    \begin{tabular}{L{2.5cm}C{1.4cm}C{1.4cm}C{1.4cm}}
        \toprule
        \multirow{2}{*}{Approach} & \multicolumn{2}{c}{DIODE}                          & \multirow{2}{*}{ScanNet} \\ \cmidrule{2-3}
                          & Indoor               & \multicolumn{1}{c}{Outdoor} &                          \\ \midrule
        LINE-MOD \cite{hinterstoisser2011gradient} & 12.839 & 17.272 & 14.479 \\
        PlanePCA \cite{jordan2014quantitative} & 10.888 & 16.579 & 13.164 \\
        3F2N \cite{fan2021three} & 10.589 & 16.254 & 12.628 \\
        SNE \cite{fan2020sne} & 10.316 & 15.431 & 12.669 \\ \midrule
        SNE+ (\textbf{Ours}) & \textbf{10.205} & \textbf{15.136} & \textbf{12.373} \\ \bottomrule
    \end{tabular}
    \label{tab.normal}
    \vspace{-1em}
\end{table}

\subsection{Evaluations for Freespace Detection}
\label{sec.freespace_detection_evaluation}
Since the whole framework shown in Fig.~\ref{fig.roadseg+} has five network levels, we can use the proposed model pruning approach to generate networks with different network depths. Considering the trade-off between accuracy and efficiency, we use the network with three levels in the rest of our experiments, and hereafter refer to it as RoadSeg+. Since we demonstrated in \cite{fan2020sne} that using surface normal information can effectively improve the freespace detection performance, we now focus on verifying the superiorities of 1) our SNE+ over SNE \cite{fan2020sne}, and 2) our RoadSeg+ over SOTA DCNNs for freespace detection. Specifically, each single-modal DCNN takes depth images as input, and each data-fusion DCNN takes RGB and depth images as input. In addition, each DCNN is evaluated with SNE \cite{fan2020sne} embedded and with our SNE+ embedded, respectively. The corresponding quantitative results are presented in Fig.~\ref{fig.road_evaluation}. It is observed that the DCNNs with our SNE+ embedded outperform themselves with SNE \cite{fan2020sne} embedded.  Moreover, our RoadSeg+ with our SNE+ embedded performs better than all other DCNNs, with an IoU increment of around 1--11\%. Qualitative results are provided in the supplement. All these results strongly prove the effectiveness of the proposed framework, which we refer to as SNE-RoadSeg+. We next submit its results to the KITTI road benchmark \cite{fritsch2013new}, as presented in Section~\ref{sec.benchmark}.

\begin{table}[t]
    \caption{Results on the KITTI road benchmark \cite{fritsch2013new}, where the best results are shown in bold type.}
    \centering
    \begin{tabular}{L{3.0cm}C{1.5cm}C{0.9cm}C{1.5cm}}
        \toprule
        Approach & MaxF~($\%$) & AP~($\%$) & Runtime~(s) \\ \midrule
        RBNet \cite{chen2017rbnet} & 94.97 & 91.49 & 0.18 \\
        LC-CRF \cite{gu2019road} & 95.68 & 88.34 & 0.18 \\
        LidCamNet \cite{caltagirone2019lidar} & 96.03 & 93.93 & 0.15 \\
        SNE-RoadSeg \cite{fan2020sne} & 96.75 & \textbf{94.07} & 0.10 \\
        PLARD \cite{chen2019progressive} & 97.03 & 94.03 & 0.16 \\ \midrule
        SNE-RoadSeg+ (\textbf{Ours}) & \textbf{97.50} & 93.98 & \textbf{0.08} \\ \bottomrule
    \end{tabular}
    \label{tab.road}
    \vspace{-1em}
\end{table}

\subsection{Performance on the KITTI Road Benchmark}
\label{sec.benchmark}
\begin{figure}[t]
    \centering
    \includegraphics[width=0.99\linewidth]{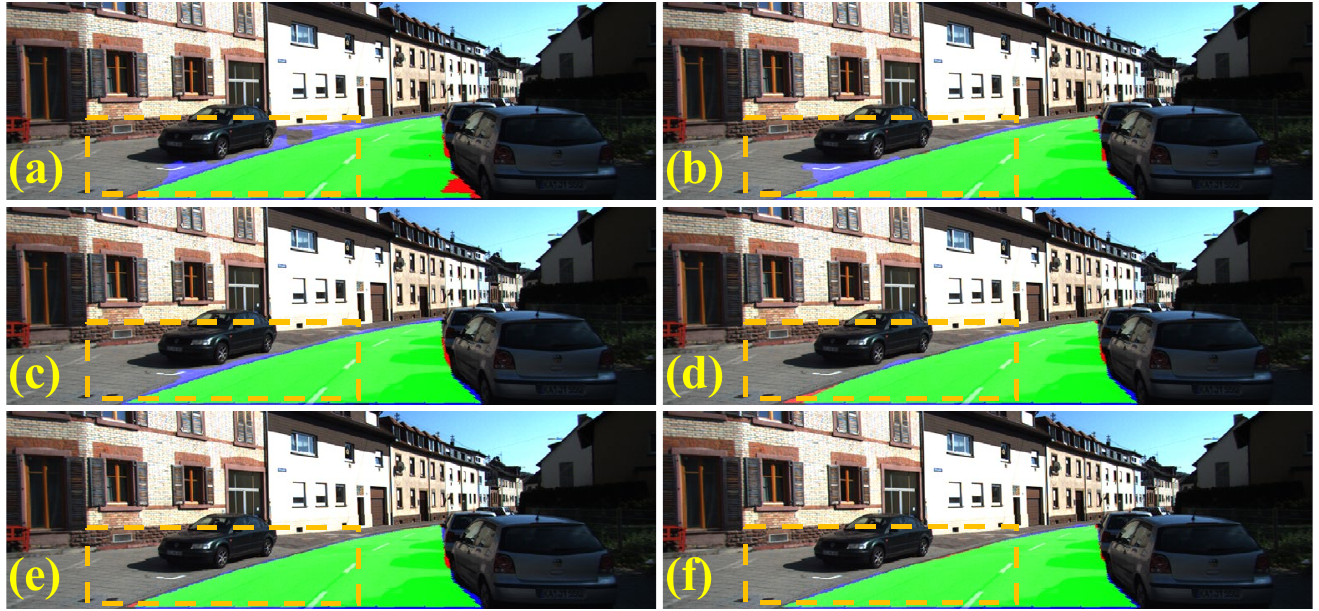}
    \caption{An example of the testing images on the KITTI road benchmark \cite{fritsch2013new}: (a) RBNet \cite{chen2017rbnet}; (b) LC-CRF \cite{gu2019road}; (c) LidCamNet \cite{caltagirone2019lidar}; (d) SNE-RoadSeg \cite{fan2020sne}; (e) PLARD~\cite{chen2019progressive}; and (f) our SNE-RoadSeg+. Green, blue and red pixels correspond to the true positives, false positives and false negatives, respectively. Significantly improved regions are marked with orange dashed boxes.}
    \label{fig.benchmark}
    \vspace{-1em}
\end{figure}

Table~\ref{tab.road} presents the KITTI road benchmark \cite{fritsch2013new} results, where our SNE-RoadSeg+ achieves the state-of-the-art performance with a real-time inference speed. Excitingly, our SNE-RoadSeg+ outperforms all other freespace detection approaches in terms of both accuracy and efficiency. Fig.~\ref{fig.benchmark} illustrates an example of the testing images on the benchmark, where we can observe that our SNE-RoadSeg+ can present more accurate freespace detection estimations. By accelerating the proposed SNE-RoadSeg+ with TensorRT, it can run in real time on resource-limited embedded computing platforms. Therefore, SNE-RoadSeg+ is more capable than SNE-RoadSeg for practical autonomous driving applications. 

\section{Conclusion}
\label{sec.conclusion}
In this paper, we proposed SNE-RoadSeg+, an effective and efficient approach for freespace detection. Our SNE-RoadSeg+ consists of 1) SNE+, a lightweight module for accurate surface normal estimation, and 2) RoadSeg+, a data-fusion DCNN that can achieve a great trade-off between accuracy and efficiency with the use of deep supervision. Extensive experimental results demonstrated 1) the effectiveness of our SNE+ for surface normal estimation, and 2) the superior performance of our SNE-RoadSeg+ over all other SOTA freespace detection approaches. Specifically, our SNE-RoadSeg+ achieves the state-of-the-art performance on the KITTI road benchmark, with a real-time inference speed.

\bibliographystyle{IEEEtran}
\bibliography{egbib}

\begin{figure*}[t]
    \centering
    \includegraphics[width=0.9\textwidth]{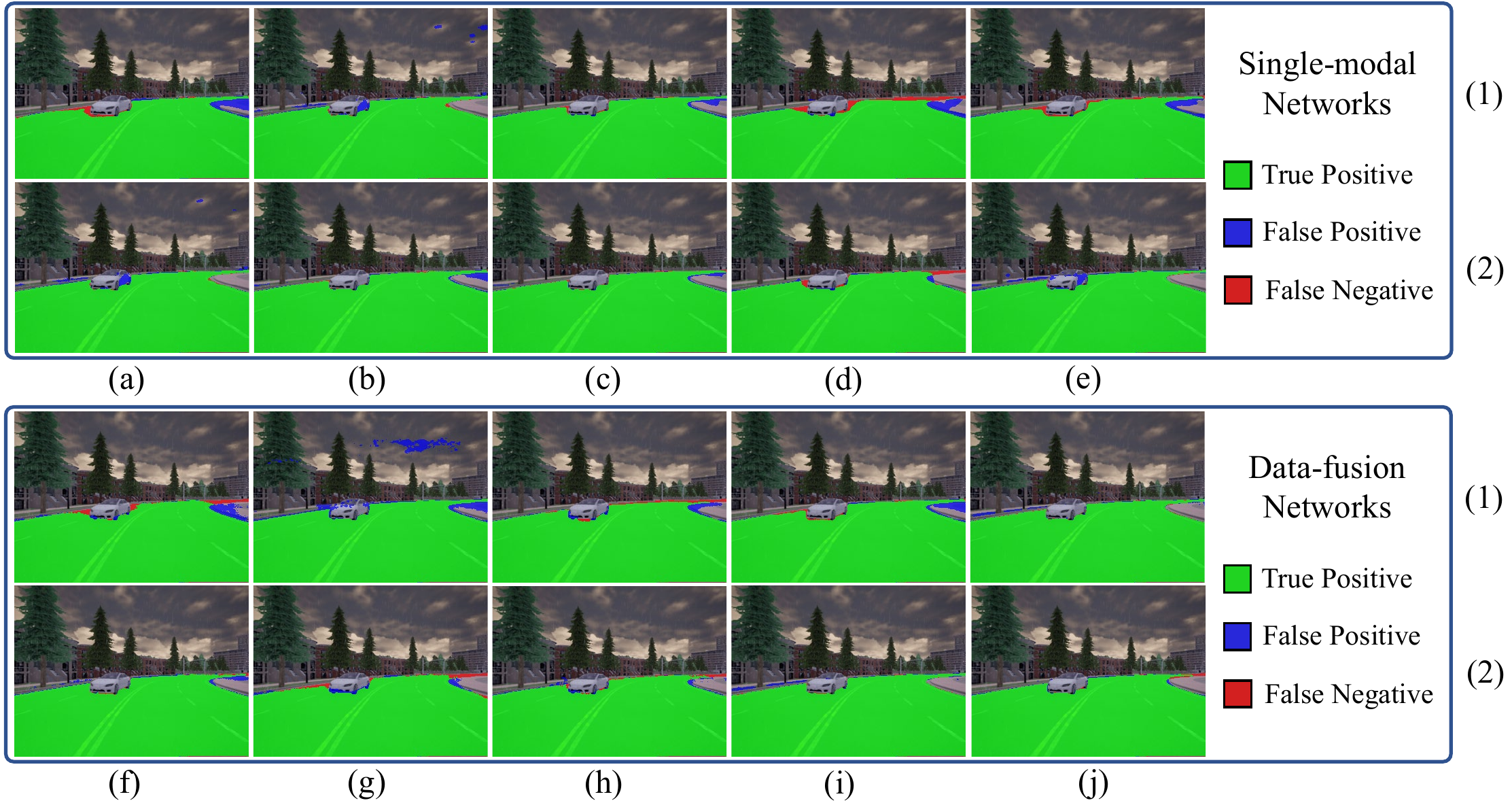}
    \caption{Examples of the freespace detection results on the R2D road dataset \cite{fan2020sne}: (a) SegNet \cite{badrinarayanan2017segnet}; (b) U-Net \cite{ronneberger2015u}; (c) DeepLabv3+ \cite{chen2018encoder}; (d) DenseASPP \cite{yang2018denseaspp}; (e) GSCNN \cite{takikawa2019gated}; (f) FuseNet \cite{hazirbas2016fusenet}; (g) Depth-aware CNN \cite{wang2018depth}; (h) RTFNet \cite{sun2019rtfnet}; (i) RoadSeg \cite{fan2020sne}; (j) our RoadSeg+; (1) DCNNs with SNE \cite{fan2020sne} embedded; and (2) DCNNs with our SNE+ embedded.}
    \label{fig.R2D}
\end{figure*}

\section*{Supplement}
\label{sec.supp}

We provide some examples of the freespace detection results in Fig. \ref{fig.R2D}, where it is evident that our RoadSeg+ with our SNE+ embedded can produce more robust and accurate freespace detection results.

In basic pinhole camera models, an observed 3D point $\mathbf{q}=(x;y;z)$ can be transformed to a 2D image pixel $\mathbf{m}=(u;v)$ using $\mathbf{m}=\mathbf{K}\mathbf{q}/z$. The local planar surface $\mathcal{S}$ of $\mathbf{q}$ satisfies:  $\mathbf{n}\mathbf{q}+d=0$, where $\mathbf{n}=(n_x;n_y;n_z)$ is the surface normal of $\mathbf{q}$ and $d$ is the distance between $\mathbf{q}$ and $\mathcal{S}$. Combining the aforementioned two equations results in:
\begin{equation}
1/z=-\Big( (u-u_o)\cdot n_x/f_x + (v-v_o)\cdot n_y/f_y +n_z
 \Big)/d,
\label{eq.sne1}
\end{equation}
where $\mathbf{m}_{o}=(u_\text{o};v_\text{o})$ is the image principal point; and $f_x$ and $f_y$ are the camera focal lengths in pixels. Differentiating (\ref{eq.sne1}) with respect to $u$ and $v$ obtains: 
\begin{equation}
    \begin{split}
        n_x=-d f_x \frac{\partial 1/z}{\partial u}, \ \  n_y=-d f_y \frac{\partial 1/z}{\partial v}.
    \end{split}
    \label{eq.nx1_ny1}
\end{equation} 
Given an arbitrary 3D point $\mathbf{p}_{i}\in \mathscr{P}$ adjacent to $\mathbf{q}$, we can obtain an $n_{z_i}$ as follows \cite{fan2020sne}:
\begin{equation}
    {n_z}_i=\frac{d}{{\Delta z_{i}}}
    \Big(
    { f_x \Delta x_{i} \frac{\partial 1/z}{\partial u} + f_y \Delta y_{i} \frac{\partial 1/z}{\partial v} }
    \Big),
    \label{eq.nz1}
\end{equation}
where $\mathbf{r}_{i}=\mathbf{p}_i-\mathbf{q}=(\Delta {x_i}; \Delta {y_i}; \Delta {z_i})$. As (\ref{eq.nx1_ny1}) and (\ref{eq.nz1}) have a common factor of $-d$, the expression of $\mathbf{n}_{i}$ (the surface normal produced by $\mathbf{q}_{i}$ and $\mathbf{p}$) is simplified as follows:
\begin{equation}
\begin{split}
\mathbf{n}_{i_j}=\Bigg(f_x \frac{\partial 1/z}{\partial u}; f_y \frac{\partial 1/z}{\partial v};-\frac{ f_x \Delta {x_i} \frac{\partial 1/z}{\partial u} + f_y \Delta {y_i} \frac{\partial 1/z}{\partial v} }{\Delta {z_i}}\Bigg).
\end{split}
\label{eq.nx2_ny2_nz2}
\end{equation}

\end{document}